%% file: main.tex
\documentclass[10pt,twocolumn,letterpaper]{article}

 \usepackage[pagenumbers]{cvpr} 

\input{preamble}

\definecolor{cvprblue}{rgb}{0.21,0.49,0.74}
\usepackage[pagebackref,breaklinks,colorlinks,citecolor=cvprblue]{hyperref}

\usepackage{multirow}
\usepackage{graphicx}
\usepackage{makecell}
\usepackage{fontawesome5}

\usepackage{amsthm}
\usepackage{graphicx}
\usepackage{color}
\usepackage{colortbl}

\usepackage{enumitem}
\usepackage[utf8]{inputenc} 
\usepackage[T1]{fontenc}    
\usepackage{url}            
\usepackage{booktabs}       
\usepackage{amsfonts}       
\usepackage{nicefrac}       
\usepackage{microtype}      
\usepackage{fontawesome5}
\usepackage{wrapfig}
\usepackage{amsmath}
\usepackage{bm}
\usepackage{bbm}
\usepackage{amssymb}
\usepackage{multirow}
\usepackage{algorithm2e}
\RestyleAlgo{ruled}
\usepackage{multirow}

\definecolor{commentcolor}{RGB}{153, 0, 0}
\definecolor{electricindigo}{rgb}{0.44, 0.0, 1.0}
\definecolor{red}{rgb}{1, 0.0, 0.0}
\definecolor{ceruleanblue}{rgb}{0.16, 0.32, 0.75}
\definecolor{fireenginered}{rgb}{0.81, 0.09, 0.13}

\usepackage{tikz}
\usepackage{pifont}
\usepackage{MnSymbol,bbding,pifont} %

\usepackage[framemethod=TikZ]{mdframed}
\definecolor{olive}{rgb}{0.6, 0.6, 0.2}
\definecolor{sand}{rgb}{0.8666666666666667, 0.8, 0.4666666666666667}
\definecolor{wine}{rgb}{0.5333333333333333, 0.13333333333333333, 0.3333333333333333}

\definecolor{deblue}{RGB}{11,132,147}
\definecolor{ocra}{RGB}{204, 119, 34}

\newcommand{\trident}{\textcolor{black}{{\fontfamily{cmtt}\selectfont TRIDENT~}}}

\newcounter{theob}[section] 
\renewcommand{\thetheob}{\arabic{theob}}

\newcounter{lemm}[section] 

\newcounter{prff}[section]

\newcounter{prf}[section]

\newcounter{propb}[section] 
\renewcommand{\thepropb}{\arabic{propb}}

\makeatletter
\renewcommand*{\@fnsymbol}[1]{\ensuremath{\ifcase#1\or \star\else\@ctrerr\fi}}
\makeatother

\title{TRIDENT: The Nonlinear Trilogy for Implicit Neural Representations}
\author{Zhenda Shen$^{1,2}$$^\star$,\: Yanqi Cheng$^{1}$\thanks{Joint first authorship.} \,, \:Raymond H. Chan$^{2}$,\: \\ Pietro Liò$^{1}$,\: Carola-Bibiane Schönlieb$^{1}$, Angelica I Aviles-Rivero$^{1}$
\smallskip
\\\:
$^{1}$ University of Cambridge, UK \,\,
$^{2}$ City University of Hong Kong, HK SAR, China \,\,
}

\begin{document}
\maketitle

\input{sec/0_abstract}

\input{sec/1_intro}

\input{sec/2_related_work}

\input{sec/3_methodology}

\input{sec/4_experiment}

\input{sec/5_ablation}

\input{sec/6_conclusion}
\input{sec/7_acknowledgement}



{
    \small
    \bibliographystyle{ieeenat_fullname}
    \bibliography{main}
}


\end{document}

%% file: preamble.tex
\usepackage[dvipsnames]{xcolor}


%% file: sec/0_abstract.tex
\begin{abstract}
Implicit neural representations (INRs) have garnered significant interest recently for their ability to model complex, high-dimensional data without explicit parameterisation. In this work, we introduce TRIDENT, a novel function for implicit neural representations characterised by a trilogy of nonlinearities. Firstly, it is designed to represent high-order features through order compactness. Secondly, TRIDENT efficiently captures frequency information, a feature called frequency compactness. Thirdly, it has the capability to represent signals or images such that most of its energy is concentrated in a limited spatial region, denoting spatial compactness. We demonstrated through extensive experiments on various inverse problems that our proposed function outperforms existing implicit neural representation functions.
\end{abstract}

%% file: sec/1_intro.tex
\section{Introduction}\label{sec:intro}

Implicit Neural Representations (INRs) have emerged as a paradigm shift, challenging the traditional grid-based data representation.INR employs a continuous coordinate-based function, typically a neural network, to encode data implicitly. This can be conceptualized as a mapping $f: \mathbb{R}^n \rightarrow \mathbb{R}^m$, where $n$ represents input dimensions (such as spatial coordinates) and $m$ denotes output dimensions (such as color and intensity in images). Unlike discrete grid representations, which are limited by their resolution, INRs excel in capturing intricate details beyond grid constraints. This attribute of INRs has been key in their adoption for a variety of inverse problems, enabling higher fidelity in detail and resolution-independent data processing.

Implicit Neural Representations (INRs) have rapidly gained prominence in diverse domains, such as medical imaging~\cite{zou2023homeomorphic}, computer graphics~\cite{mildenhall2021nerf}, and robotics~\cite{khargonkar2023neuralgrasps}, due to their remarkable practical applicability. INRs offer a unique approach to encoding complex data, leveraging deep neural networks to represent information implicitly and continuously. Researchers have undertaken substantial efforts to improve INRs, introducing various techniques from sinusoidal functions like SIREN to wavelet transforms like WIRE~\cite{saragadam2023wire}. Despite these advancements, existing INR techniques grapple with critical challenges.

Firstly, a fundamental issue arises from the specialised nature of existing INR techniques, which are often  designed for specific tasks, resulting in a lack of adaptability and generalisability across diverse problem domains[ref]. This inherent task-specific imposes limitations on their broader applicability, constraining their efficacy in addressing a wide spectrum of inverse problems.

Secondly, current INR functions have inherent difficulties when attempting to capture intricate fine-grained details while simultaneously avoiding the introduction of undesirable artifacts. Notably, they often encounter challenges in representing high-order information within low-order domains, which leads to a degradation of information fidelity and the generation of inaccurate representations. Furthermore, these methods may exhibit suboptimal performance in efficiently encoding frequency information and spatial regions, thereby impeding their effectiveness in representing complex data structures in several tasks.

In this study, we introduce a novel Implicit Neural Representation function named \trident. Our approach is distinguished by a trilogy of nonlinearities, which set it apart from existing techniques in the field.
Firstly, we incorporate order compactness into \trident, enabling the transformation of high-order information into a low-order representation. This part enhances our function's capacity to represent complex data efficiently. Secondly, frequency compactness is another component of our trilogy, boosting \trident to adeptly capture and represent frequency information, ultimately improving its fidelity in encoding intricate data structures.
Lastly, our function offers spatial compactness, ensuring that it represents signals or images with a focus on concentrating the majority of their energy within limited spatial regions. This feature enhances \trident's ability to faithfully capture essential spatial features. Our contributions are summarised next.

\begin{itemize}
    \item[] \faHandPointRight We introduce a novel function called \trident for implicit neural representation, that stands out for:
    \begin{itemize}
        \item \trident proposed a unique trilogy of nonlinearities being the first INR function that carefully is designed to enforce better approximations that generalise well across several inverse problems.
        \item \trident proposed trilogy enforces: higher-order features (order compactness), better representation of frequency information (frequency compactness), and concentration on useful spatial regions for fine-grained details (spatial compactness).
    \end{itemize}
    \item[] \faHandPointRight  We demonstrate the capabilities of our \trident function through extensive experiments across various inverse problems, showing that it not only outperforms existing methods in the literature but also generalises well across tasks, consistently delivering the best performance
\end{itemize}

%% file: sec/2_related_work.tex
\section{Related Work}\label{sec:work}
This section revisits existing literature on Implicit Neural Representations and the core principles of our novel technique, providing a basis for contrasting it with current solutions.

\smallskip
\textbf{Implicit Neural Representation.} In contrast to conventional grid-based approaches, Implicit Neural Representations (INR), built upon multilayer perceptron (MLP) networks, offer a continuous and differentiable framework for signal recovery~\cite{Yuce_2022_CVPR}. Previous research has demonstrated the versatility of INR in addressing a wide spectrum of challenges across diverse domains, ranging from medical imaging~\cite{zou2023homeomorphic}, signal processing~\cite{sitzmann2020implicit}, occupancy volume~\cite{mescheder2019occupancy} to computer graphics~\cite{mildenhall2021nerf}.

The community has explored different alternatives to improve INR. The traditionally popular ReLU activation function fails to represent features in the high-frequency domain well~\cite{hao2022implicit}, leading to low accuracy of ReLU-INR. Several improvements in network architectures have been proposed in recent years.  Borrowing the idea from position encoding~\cite{muller2022instant,mildenhall2021nerf}, the authors of that~\cite{tancik2020fourier} proposed Fourier Encoding to represent high-frequency features in low-dimensional domains. Despite a significant improvement in representativeness, training with the ReLU activation function still results in low accuracy. 

Inspired by the principles of Fourier Transformation, the authors introduced a network leveraging sinusoidal functions, referred to as SIREN~\cite{sitzmann2020implicit}, with the aim of enhancing the representational capacity of the network. Additionally, it is noteworthy that various other alternatives to nonlinearity have been under consideration including~\cite{ramasinghe2022beyond}. The Multiplicative Filter Networks, use sinusoidal functions and the Gabor wavelet to function as a filter to replace the activation function in INR architecture~\cite{fathony2020multiplicative} Analogous to SIREN, the WIRE model, as proposed in~\cite{saragadam2023wire}, also adopts the Gabor wavelet as the activation function for the Implicit Neural Representation (INR).

\medskip
\textbf{Inverse problems.} Inverse problems manifest in several crucial practical applications, spinning from image restoration to acoustic source reconstruction. The community has discussed rigorous mathematical foundations~\cite{devaney2012mathematical, bertero2021introduction, vogel2002computational} for understanding inverse problems on imaging, from its ill-posedness~\cite{rudin1992nonlinear, candes2006robust, venkatakrishnan2013plug} to the parameter selection~\cite{wei2022tfpnp}.

The classical approaches for addressing inverse problems are knowledge-driven. One category involves providing approximating solutions with boundary conditions that guarantee the existence of the ill-posed problems~\cite{engl1996regularization, kirsch2011introduction}, while the other integrates knowledge of targeted parameter structures with sparsity assumptions~\cite{daubechies2004iterative, jin2012sparsity} or stochastic models~\cite{kaipio2007statistical, mueller2012linear}. However, the classical methods have their inherent limitations. The former approaches suffers from limited physical priors from the observation and high computational cost for accurate analytical models, the latter is limited by its instability in capturing the task-specific data structures~\cite{arridge2019solving}.

Recent advancement in deep learning techniques also offers solutions on inverse problems with data-driven models~\cite{ongie2020deep, mccann2017convolutional}. They take data examples on specific problems with the prior knowledge which generates the data for training to learn the data structure, while in many applications the lack of sufficient data prevents it from being robust, thus insufficient to support an entirely data-driven approach. A recent avenue of progress is now combining the knowledge- and data-driven models to solve inverse problems. Several algorithms~\cite{venkatakrishnan2013plug,  lucas2018using} have been proposed for this combination, yet these findings predominantly remain experimental and lack a comprehensive theoretical underpinning. Nevertheless, there is an emergence of certain mathematical concepts for addressing data-driven approaches in the context of inverse problems~\cite{arridge2019solving}. 
The majority of data-driven techniques rely on a large amount of data, hindering their wide applicability. Implicit Neural Representations (INR) allow solving and deriving inverse problems from a single sample.

%% file: sec/3_methodology.tex
\section{Methodology}\label{sec:method}
This section details two key aspects of our \trident approach: firstly, the formulation, and secondly, its theoretical foundation.

\subsection{TRIDENT Formulation}
In this work, we introduce a novel implicit neural representation function, which stands out for its several advantages over existing methodologies in the field. Our new function reads:
\begin{equation} \label{eq:main}
    \vspace{-0.2cm}\exp(-|s_0(\left[x, \cdots \cos \left(2 \pi \sigma^{j / m} \mathbf{x}\right), \sin \left(2 \pi \sigma^{j / m} \mathbf{x}\right)\cdots \right]^{\mathrm{T}}) |^2),
\end{equation}
%
where  $\sigma$ represents the frequency parameter and $m$ is the mapping size, while $s_0$ denotes the scale parameter. 
Our function~\eqref{eq:main} is then considered within our defined network architecture, which is structured as follows: 
\begin{equation}
    \Phi(\mathbf{x})=\mathbf{W}_{n}\left(\phi_{n-1} \circ \phi_{n-2} \circ \ldots \circ \phi_{1}\circ \gamma\right)(\mathbf{x})+\mathbf{b}_{n}, 
\end{equation}
where each element of the input vector $\mathbf{x}_{i}$ is transformed by $\phi_{i}$ as:
$$\mathbf{x}_{i} \mapsto \phi_{i}\left(\mathbf{x}_{i}\right)=\exp \left(-s_0|\mathbf{W}_{i} \mathbf{x}_{i}+\mathbf{b}_{i}\right|^{2}).$$
Moreover, the first-layer mapping, $\gamma$, is defined as:
$\gamma(\mathbf{v})=\left[\mathbf{v}, \ldots, \cos \left(2 \pi \sigma^{j / m} \mathbf{v}\right), \sin \left(2 \pi \sigma^{j / m} \mathbf{v}\right), \ldots\right]^{\mathrm{T}},$
where $\phi_{i}: \mathbb{R}^{M_{i}} \mapsto \mathbb{R}^{N_{i}}$ is the $i^{th}$  layer of the network.

\begin{figure*}[t!]
    \centering
    \includegraphics[width=1\linewidth]{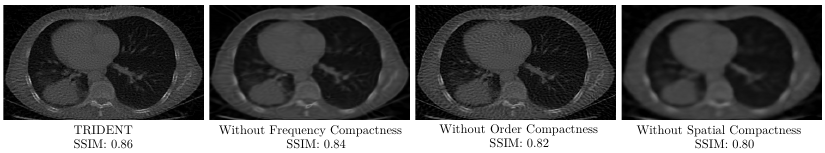}
     \caption{Visualisation comparison of CT reconstruction task on 100 projections among \trident, and \trident without Frequency, Order, and Spatial Compactness separately.} 
    \label{fig:ctteasor}  
\end{figure*}

\subsection{Theoretical Foundation of TRIDENT}

\textit{What properties characterise our proposed implicit neural representation function, and how do they set it apart from existing models?} Unique in its approach, our function is distinguished by a set of theoretically-motivated properties. These properties not only endow our function with novel capabilities but also align it with fundamental theoretical principles that have not been fully explored in prior work. Without losing generality, we simplify  \eqref{eq:main} to $\exp \left(-| \cos (x)|^{2}\right)$ and let $\phi(x)=\phi_1 \circ \gamma (x)$.
We next introduce \trident foundation based on a trilogy of nonlinearities.

\smallskip

\textcolor{wine}{\faHandPointRight[regular] \textbf{Order Compactness.}}
In real context, a continuous function on closed intervals can be expanded as a series of cosine functions, e.g., $\sum_{n=0}^{\infty } A_n \cos \frac{n \pi x}{l} $. Meanwhile, known from the Stone–Weierstrass theorem, continuous functions on closed intervals can be uniformly approximated by polynomial series. Inspired by the above two theorems, we aim to enable our function, $\phi(x)$ to represent high-order information in the low-order domain, augmenting the representational capacity of our function. By elevating the series index $n$ to the exponent of the cosine function in the Fourier cosine series, we obtain:
\begin{equation}
    \phi (x)=\sum_{n=0}^{\infty } A_n \cos^n \Big(\frac{\alpha \pi x}{l}\Big),
\label{ourformulasimpl}
\end{equation}
where $\alpha$ is the frequency coefficient and $l$ is the length of the domain. Without losing of generality, let $\alpha = \frac{2}{\pi}$ and $l = 1$, yielding: $\phi (x)=\sum_{n=0}^{\infty } A_n \cos^n (2 x)$.\\
Consequently, with  $\phi(x)$ as our formula, the network is enabled to represent high-order features, which we call it \textit{Order Compactness}.\\
\textcolor{wine}{\faHandPointRight[regular] \textbf{Frequency Compactness.}} If a continuous function is expanded as a series of cosine functions, each appropriately scaled and shifted, it implies that this Fourier series provides an efficient way to represent frequency features, a concept we can refer to as frequency compactness. Consequently, if our formula takes the form as in~\eqref{ourformulasimpl}, it will inherently preserve the strong frequency compactness characteristic of the Fourier cosine series, as demonstrated by:
\begin{equation}
\label{frequency}
    \cos ^{n} \theta=\frac{1}{2^{n}}\left(\begin{array}{c}
n \\
\frac{n}{2}
\end{array}\right)+\frac{2}{2^{n}} \sum_{k=0}^{\frac{n}{2}-1}\left(\begin{array}{l}
n \\
k
\end{array}\right) \cos [(n-2 k) \theta]
\end{equation}
which means that our formula, $\phi(x)$, like the property of the Fourier series, has the capacity to represent frequency information efficiently, bringing our method frequency compactness.

\textcolor{wine}{\faHandPointRight[regular] \textbf{Spatial Compactness.}} Apart from the frequency compactness and the spatial compactness already brought by $\phi(x)$ in the form of~\eqref{ourformulasimpl}, we aim to utilise the choice of $A_n$ to further enhance spatial compactness.  This refers to the ability to represent a signal or an image in a way that concentrates most of its energy in a limited spatial region.
If we define $\alpha_n$ as $\alpha_n=\frac{(-1)^{n+j}}{2^{n+j} \cdot(n !) \cdot(j !)}$. Given that $|\alpha_n| \leq \frac{1}{2^{n+j}}$, and  consider the convergence of $\sum_{j=0}^{\infty} \frac{1}{2^{n+j}}$ to $\frac{1}{2^n}$, the series  $\sum_{j=0}^{\infty} \frac{(-1)^{n+j}}{2^{n+j} \cdot(n !) \cdot(j !)}$ is determined to converge uniformly by the Weierstrass M-test. Consequently, this allows us to express $A_n$ as: 
$$A_n=\sum_{j=0}^{\infty} \frac{(-1)^{n+j}}{2^{n+j} \cdot(n !) \cdot(j !)}.$$
We then get:
\begin{equation}
\begin{aligned}
\phi(x)=& \sum_{n=0}^{\infty}\left(\sum_{j=0}^{\infty} \frac{(-1)^{n+j}}{2^{n+j} \cdot(n !) \cdot(j !)}\right) \cos ^{n}(2 x)\\
    = & \sum_{n=0}^{\infty} \frac{(-1)^{n}}{2^{n} n !} \cos ^{n}(2 x)+\frac{(-1)^{n+1}}{2^{n+1} n !} \cos ^{n}(2 x)\\ 
    &+\frac{(-1)^{n+2}}{2^{n+2} n ! 2!} \cos ^{n}(2 x) +\frac{(-1)^{n+3}}{2^{n+3} n ! 3 !} \cos ^{n}(2 x)+\cdots \\
    = & \sum_{i=0}^{\infty} \frac{(-1)^i}{2^{i} i !}\left[\cos ^{i}(2 x)+\left(\begin{array}{l}
i \\
1
\end{array}\right){\left.\cos^{i-1}(2 x)+\cdots\right]}\right. \\
=&  \sum_{i=0}^{\infty} \frac{(-1)^i}{2^{i} i !}(1+\cos (2 x))^i \\
=&  \sum_{i=0}^{\infty} \frac{(-1)^i}{i !}\left(\frac{1+\cos (2 x)}{2}\right)^{i} \\
=&  \sum_{i=0}^{\infty} \frac{(-1)^i\cos^{2i} \left(x\right.)}{i !} \\
\end{aligned}
\end{equation}
Linking back to the exponential series:  $\exp{x}=\sum_{i=0}^{\infty} \frac{x^i}{i!}$,
we finally get: 
\begin{equation}
    \phi (x)=\sum_{n=0}^{\infty } A_n \cos^n (2 x)=\exp \left(-| \cos (x)|^{2}\right)
\end{equation}
which becomes our simplified formula. Thanks to the Gaussian window, we can represent the signal or image in
a well-localised in the spatial domain. Consequently, our simplified formula enjoys the benefit of the nonlinear Trilogy: order, frequency, and spatial compactness.

However, it is well known that while the full Fourier cosine series effectively represents the even part of a function, it falls short of capturing the odd part. Our approach enhances the representativeness of odd components by incorporating dedicated sine channels. Additionally, to ensure the preservation of non-Fourier components, an input channel is integrated into our system, with which comes the three-channel formula: $\exp(-|s_0(\left[x,  \cos \left(2 \pi  \mathbf{x}\right), \sin \left(2 \pi  \mathbf{x}\right) \right]^{\mathrm{T}}) |^2)$. Finally, for a more efficient representation, we introduce a frequency parameter $\sigma$ and apply the "log-linear" strategy, resulting in our final function:
$$\exp(-|s_0(\left[x, \cdots \cos \left(2 \pi \sigma^{j / m} \mathbf{x}\right), \sin \left(2 \pi \sigma^{j / m} \mathbf{x}\right)\cdots \right]^{\mathrm{T}}) |^2)$$

We demonstrate the nonlinearity trilogy of our formula, which encompasses order, frequency, and spatial compactness. To visually illustrate the impact of each compactness type, we refer to Figure~\ref{fig:ctteasor}. The outputs in this figure were generated using the setting described in Section~\ref{ct}, using 100 projections. \trident achieved an SSIM of 0.86. Removing the sinusoid channels reduced the SSIM to 0.84, impacting sensitivity to high-frequency features and detail preservation. Using WIRE~\cite{saragadam2023wire}, we demonstrated the loss of high-order information without order compactness. Substituting the Gaussian window with ReLU activation, to test the absence of spatial compactness, resulted in an over-smoothed output with an SSIM of 0.80. These results underscore TRIDENT's advantage in representing both odd and even features efficiently.

%% file: sec/4_experiment.tex
\section{Experiment}\label{sec:experiment}
This section details all experiments conducted to validate our proposed \trident function.

\subsection{Implementation Details \& Evaluation Protocol}
For fair comparison in all experiments, we used a share code-base. Every experiment was conducted using the PyTorch platform and the Adam optimiser~\cite{kinga2015method}.  In this work, we consider five tasks: image denoising, 3D occupancy reconstruction, image super-resolution, audio reconstruction, and CT reconstruction.
We evaluate each task using its corresponding metric, as stated in the respective task descriptions

If not otherwise mentioned, we use 2 hidden layers with 256 hidden features in each layer. Apart from the Audio task, we all added the scheduler to reduce the learning rate during training. 
We compared our function against the current literature including Positional Encoding~\cite{mildenhall2021nerf}, MFN~\cite{fathony2020multiplicative}, SIREN~\cite{sitzmann2020implicit}  and WIRE~\cite{saragadam2023wire}. 
For SIREN nonlinearity, we set the frequency parameter as 10 as suggested in~\cite{sitzmann2020implicit}, and use the recommended parameters for WIRE from~\cite{saragadam2023wire}. 
\input{tab/experimenttable}

\subsection{Image Denoising}
For evaluating our method's robustness, we reconstruct a colourful (with high-frequency features) image from the DIV2K dataset~\cite{agustsson2017ntire}. For orthogonality to the Gaussian window, we chose to add independently poison-distributed noise to each pixel, with a maximum average photon noise of 30 and an integration time of 2.

\begin{figure*}[t!]
    \centering
    \includegraphics[width=1\linewidth]{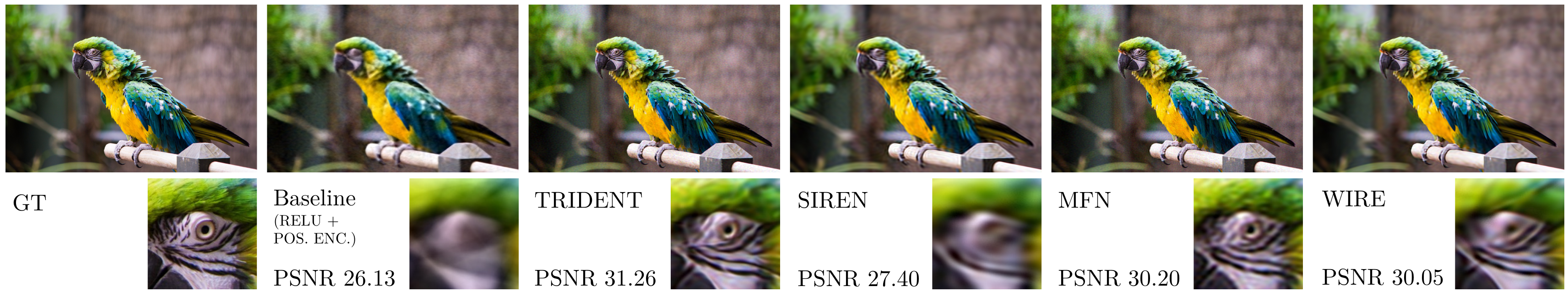}
     \caption{The visualization comparison of the denoising task on the "Parrot" example among the Baseline (ReLU + Position Encoding), OURS, SIREN, MFN, and WIRE methods.}
    \label{fig:denoising}  
\end{figure*}

We use the Peak Signal-to-Noise Ratio (PSNR) as the metric to evaluate the experiment and train with 2K iterations with different nonlinearities. We use $l_2$ loss between the INR output and the input image. Figure~\ref{fig:denoising} shows the denoising result of each nonlinearity. 

\begin{figure*}[t!]
    \centering
    \includegraphics[width=1\linewidth]{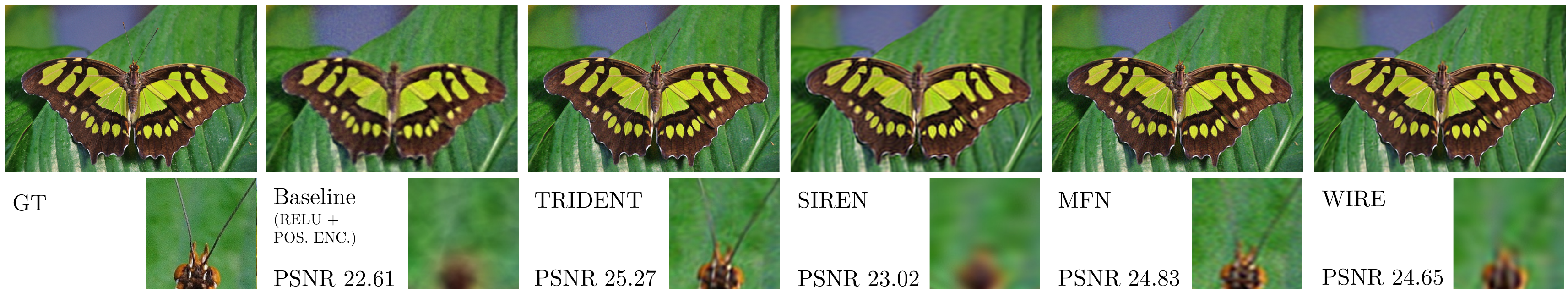}
     \caption{Visual comparison of the denoising task on the "Butterfly" example among the Baseline (ReLU + Position Encoding), OURS, SIREN, MFN, and WIRE methods.} 
    \label{fig:denoisingbutterfly}  
\end{figure*}

Figures \ref{fig:denoising}\&\ref{fig:denoisingbutterfly} show the denoising result for the parrot and butterfly images. Our method not only achieves the best performance in both cases but is also the only one that effectively preserves detailed features, demonstrating its robustness to noise.  MFN and WIRE rank as the second-best, generally recovering the images but falling short in detail preservation.  SIREN and the Baseline (ReLU+POS.ENC.) perform the least effectively, often resulting in over-smoothed outcomes. These examples underscore our method's robustness to noise, attributable to its spatial compactness, which is crucial for capturing important information concentrated in specific image regions.
\subsection{3D Occupancy Reconstruction}
To evaluate our method's representativeness in a higher dimension, we learn to represent the occupancy volume of a 3D image, the Thai statue. We sample the 3D image into a $512^3$ grid, assigning  a value of 1 to pixels outside the 3D shape and 0 to those inside.  For the marching cubes algorithm, we set the threshold at 0.5. The Intersection over Union (IoU) value was used as a metric to measure the reconstruction level. During the experiments, we employed the Mean Squared Error (MSE) as the loss function and trained the model with 4 different nonlinearities for 300 iterations.

\begin{figure*}[t!]
    \centering
    \includegraphics[width=1\linewidth]{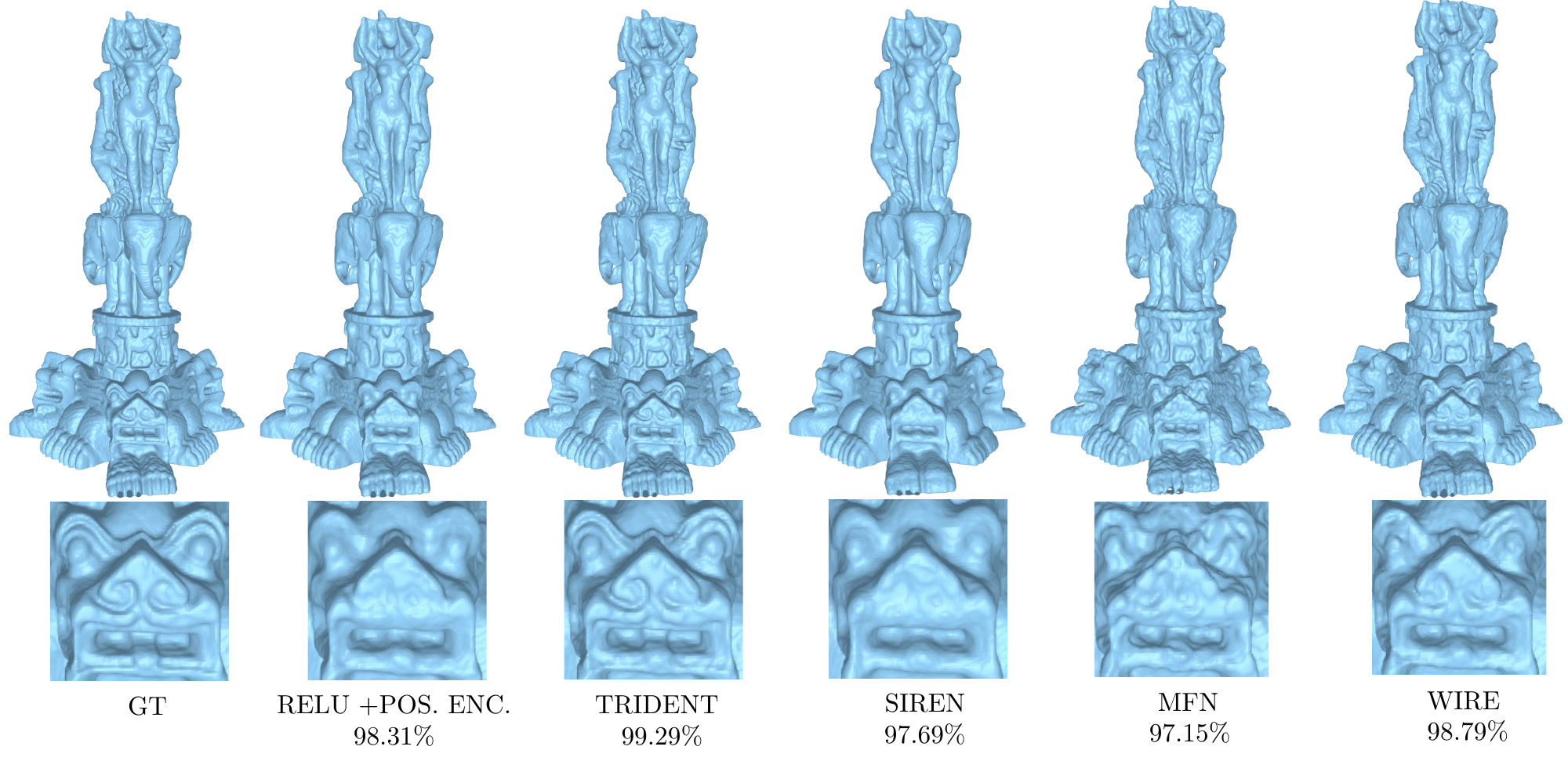}
    \caption{The visualisation comparison of 3D Occupancy Reconstruction (4 $\times$) task on "Thai-Statue" example among the Baseline (ReLU + Position Encoding), OURS, SIREN, MFN, and WIRE methods.} 
    \label{fig: 3d}  
\end{figure*}

Figure~\ref{fig: 3d} visualises the results of the  3D Occupancy Reconstruction. \trident achieves a 99.29\% score, nearly perfect,  successfully reconstructing all fine details of the 3D Thai statue input. WIRE and the baseline (ReLU+ POS. ENC.) attain scores between 98\% and 99\%, but lose some details during training. SIREN and MFN, with results ranging between 97\% and 98\%, demonstrate less representational efficacy as the dimensionality increases. This experiment highlights \trident's superior representativeness not only for low-order features but also for high-order features. \\

\begin{figure*}[t!]
    \centering
    \includegraphics[width=1\linewidth]{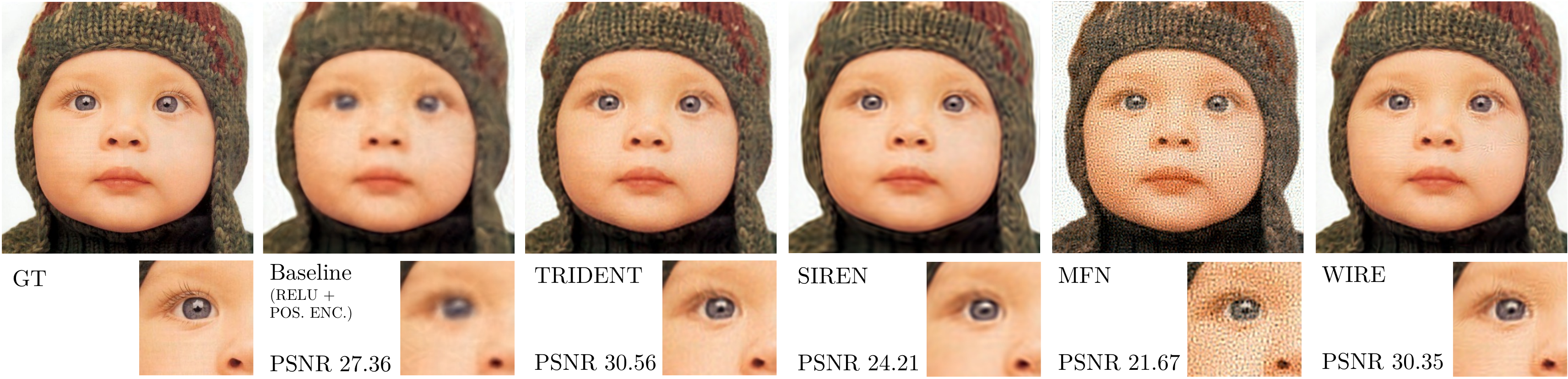}
    \caption{Visual comparison of single image super-resolution (4 $\times$) task on "Baby" example among the Baseline (ReLU + Position Encoding), OURS, SIREN, MFN, and WIRE methods.} 
    \label{fig: SISRbaby}  
\end{figure*}

\begin{figure*}[t!]
    \centering
    \includegraphics[width=1\linewidth]{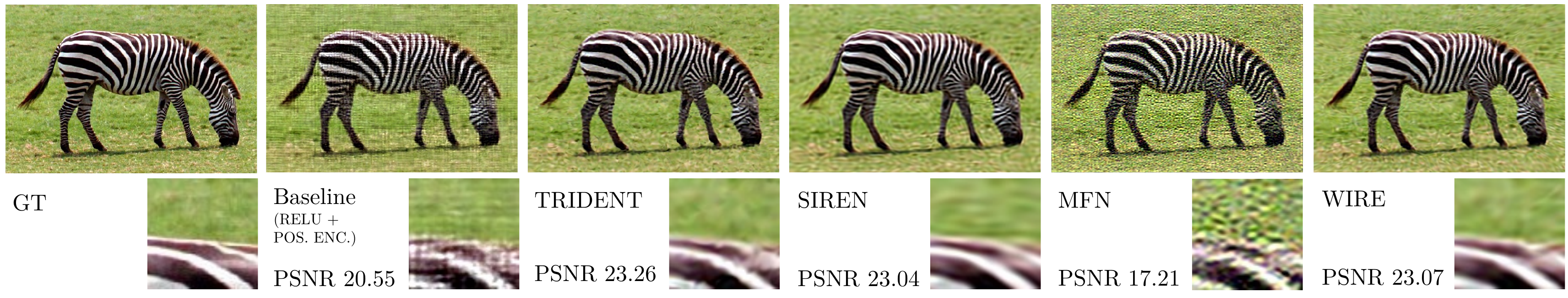}
     \caption{Visualisation comparison of single image super-resolution (4 $\times$) task on "Zebra" example among the Baseline (RELU + Position Encoding), OURS, SIREN, MFN, and WIRE methods.} 
    \label{fig: SISRzebra}  
\end{figure*}
\subsection{\label{sr}Single Image Super Resolution}
In the task of image super-resolution, we aim to construct a high-resolution image from a low-resolution one. INR's high representativeness is particularly beneficial for super-resolution. 
We used the well-known datasets Set5~\cite{bevilacqua2012low} and Set14~\cite{zeyde2012single}. We implemented 4X super-resolution on four images randomly selected from these two datasets. For each task, we trained the model for 2000 iterations, using the
$l_2$ loss between the input low-resolution picture and the INR output. We also employed the Peak Signal-to-Noise Ratio (PSNR) between the INR output and the ground truth to assess the quality of the images. Our method outperformed the other nonlinearities in all four tasks.
Table~\ref{fig: SISRbaby} presents the 4X super-resolution results for `the baby' image using different nonlinearities, and Figure~\ref{fig: SISRbaby} visualises these results.

TRIDENT achieved the highest average PSNR (26.61 dB) and produced a smooth result with no obvious noise. WIRE and SIREN almost completely recovered the original images, albeit with some colour striation in the output. 
The baseline (ReLU+ POS. ENC.) was unable to fully recover the `zebra' sample but performed relatively well with the 'baby' image. However, MFN showed the least effective performance in both images. Across these tasks, TRIDENT demonstrated its strong frequency compactness by effectively recovering both high-frequency features (as in the zebra task) and low-frequency features (as in the baby task) compared to the others, showcasing its stability and robustness during training to avoid colour striations.
\begin{figure*}[t!]
    \centering
    \includegraphics[width=1\linewidth]{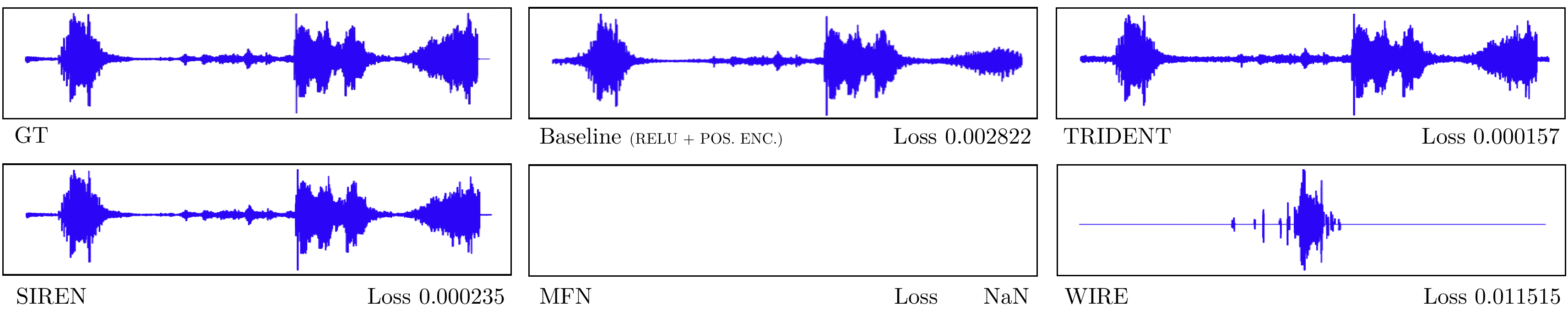}
     \caption{Visual comparison of audio reconstruction task on "Friends" example among the Baseline (ReLU + Position Encoding), OURS, SIREN, MFN and WIRE methods.} 
    \label{fig: audiofriends}  
\end{figure*}
\begin{figure*}[t!]
    \centering
    \includegraphics[width=1\linewidth]{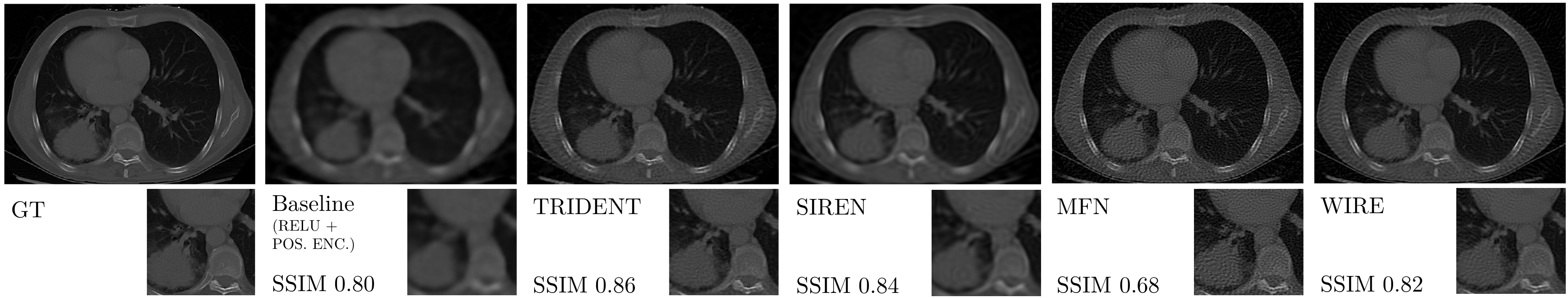}
     \caption{Visualisation comparison of CT reconstruction task on 100 projections among the Baseline (ReLU + Position Encoding), OURS, SIREN, MFN and WIRE methods.} 
    \label{fig:ct}  
\end{figure*}
\subsection{\label{audio}Audio Reconstruction}
Neural networks have shown the capability to represent audio signals~\cite{agustsson2017ntire}. However, it is hard for an MLP to solve the inverse problem of audio because it contains many kinds of nonlinearities including order and frequency features. To showcase the high representativeness and multimodality of our method compared to others, we aim to represent audio signals using our INR model. We evaluated the performance of MLPs with different nonlinearities on a selected four-second audio clip from the TV show `Friends'. This clip features not only Joey's speech but also the background laughter of the audience. The model was trained using the Mean Squared Error (MSE) loss between the INR output and the original audio over 9000 iterations. We employ a 5-layer MLP with 256 hidden units in each layer for this task.

Figure~\ref{fig: audiofriends} shows the visualisation of the audio signal we successfully recovered. Apart from SIREN and ours, other INR approaches failed to reconstruct the waveforms accurately. Pos+ReLU can represent the speech of Joey's speech but failed to capture the audience's laughter (particularly in the final $\frac{1}{4}$ of the waveform). WIRE resulted in an incorrect audio file, and MFN was unable to construct a waveform due to issues with gradient descent.   Only our model and SIREN were able to accurately reconstruct the audio signal, with our model achieving a lower loss than SIREN. This experiment highlights the exceptional representativeness of our method, benefiting from the \textit{Nonlinear Trilogy}, where order compactness and frequency compactness aid in detailed recovery, and spatial compactness on specific regions of the signal.

\begin{figure*}[t!]
    \centering
    \includegraphics[width=1\linewidth]{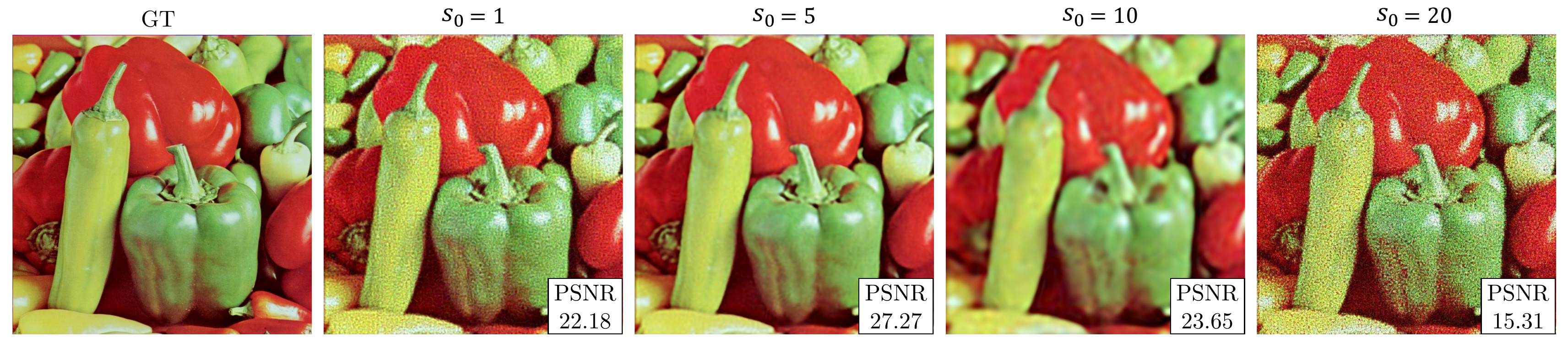}
     \caption{The 4X super-resolution result for TRIDENT with $s0=1,5,10,20$ separately. } 
    \label{fig:ablation_parameter}  
\end{figure*}
\begin{figure*}[t!]
    \centering
    \includegraphics[width=\linewidth]{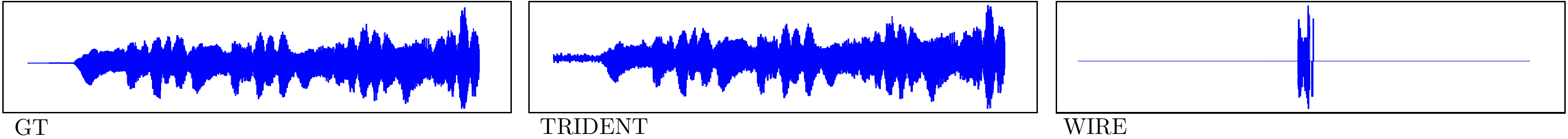}
     \caption{The visualisation of audio wave on a musical example "Bach’s Cello Suite No.1" after reconstruction by OURS and WIRE methods.} 
    \label{fig:ablation_audio}  
\end{figure*}

\begin{figure}[t!]
    \centering
    \includegraphics[width=\linewidth]{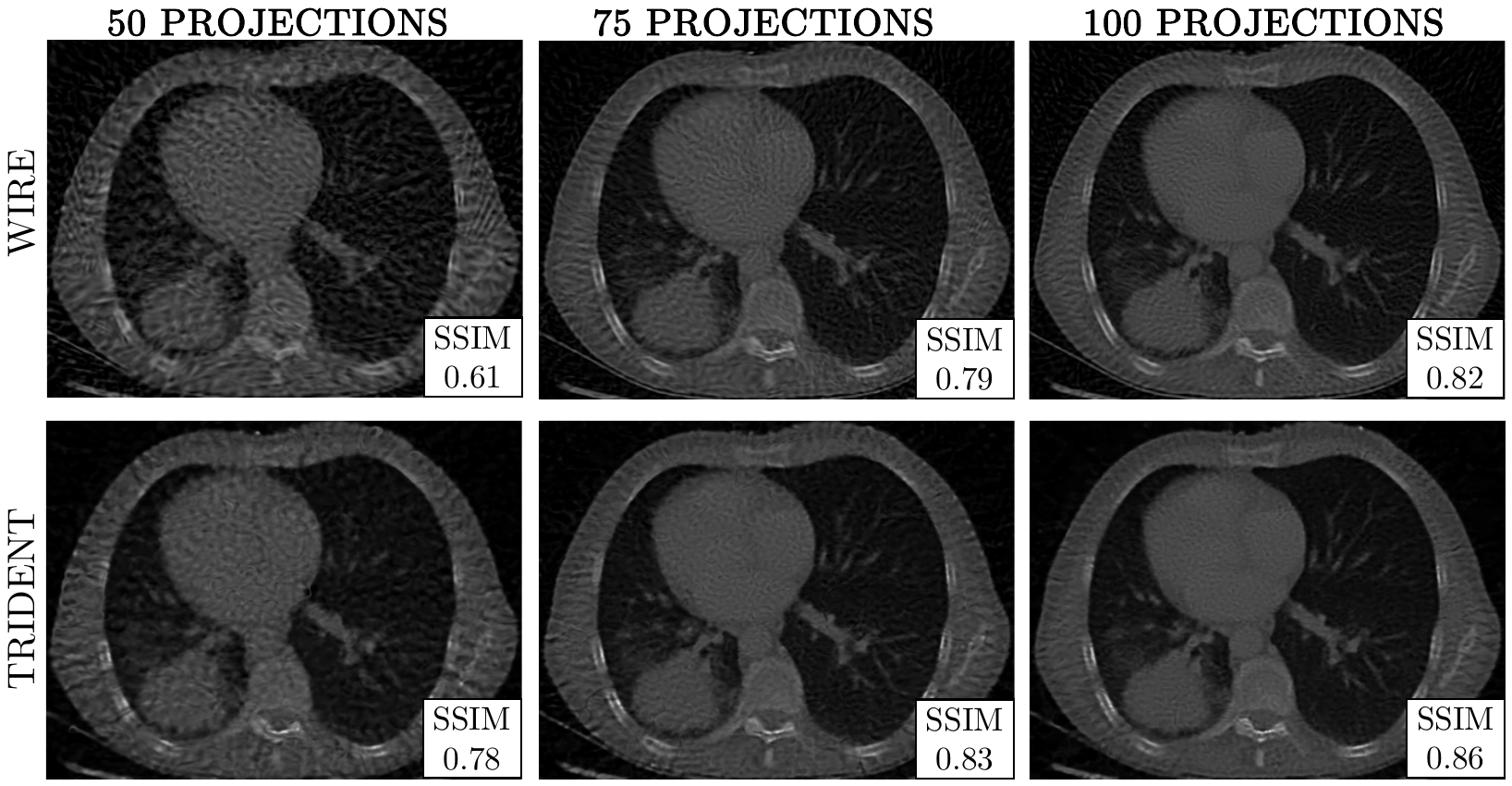}
     \caption{The CT reconstruction with 1, 5, 10 and 20 projections} 
    \label{fig:ablation_CT}  
\end{figure}

\subsection{\label{ct}CT reconstruction}
To further test our method's efficacy in solving underconstrained problems with priors, we conducted an experiment focused on reconstructing a  CT image of a $256^2$ x-ray chest image~\cite{clark2013cancer}, a challenging task for MLPs due to the presence of both even and odd features in the image. In the main experiment, we reconstructed the image with 100 CT measurements. We increase the hidden features from 256 to 300 with 2 hidden layers and train with 5000 iterations. SSIM~\cite{wang2004image} between the ground truth and the INR output served as the metric to measure the restoration quality. 

Figure~\ref{fig:ct} displays the visualizations of the results from different nonlinearities. Our method produces the sharpest image with clearly defined features.  SIREN and WIRE also yield clear outputs, but their results include some striation artifacts. The baseline approach (ReLU +POS.ENC.) leads to a blurred image. MFN performed the worst in this task, resulting in an output image with excessive noise. This demonstrates that our method, incorporating both odd and even elements in its formula, serves as an effective representation tool for solving underconstrained inverse problems.

%% file: tab/experimenttable.tex



\begin{table*}[]
\centering
\resizebox{1\textwidth}{!}{%
\begin{tabular}{lccccccccc}
\hline
\multicolumn{1}{l|}{\multirow{2}{*}{\textsc{Method}}} & \multicolumn{2}{c|}{\cellcolor[HTML]{EFEFEF}Denoising (PSNR)} & \multicolumn{1}{c|}{\cellcolor[HTML]{EFEFEF}Occupancy (IoU)} & \multicolumn{4}{c|}{\cellcolor[HTML]{EFEFEF}Super resolution (PSNR)} & \multicolumn{1}{c|}{\cellcolor[HTML]{EFEFEF}Audio (Loss)} & \multicolumn{1}{c}{\cellcolor[HTML]{EFEFEF}CT Recon (SSIM)} \\ \cline{2-10} 
\multicolumn{1}{l|}{} & Parrot & \multicolumn{1}{c|}{Butterfly} & \multicolumn{1}{c|}{Thai-Statue} & Zebra & Baby & Monarch & \multicolumn{1}{c|}{Pepper} & \multicolumn{1}{c|}{Friends} & 100 projections \\ \hline
SIREN~\cite{sitzmann2020implicit} & 27.40 & 23.02 & 97.69  \% & 23.04 & 28.99 & 24.21 & 26.63 & 0.000235 & 0.84 \\
WIRE~\cite{saragadam2023wire} & 30.05 & 24.65 & 98.79  \% & 23.07 & 30.35 & 25.18 & 27.21 & 0.011515 & 0.82 \\
RELU+ POS.ENC.~\cite{mildenhall2021nerf} & 26.13 & 22.61 & 98.31\% & 20.55 & 27.36 & 23.84 & 24.83 & 0.002822 & 0.80 \\
MFN~\cite{fathony2020multiplicative} & 30.20 & 24.83 & 97.15\% & 17.21 & 21.67 & 21.51 & 20.39 & NaN & 0.68 \\
TRIDENT & \cellcolor[HTML]{D7FFD7}\textbf{31.26} & \cellcolor[HTML]{D7FFD7}\textbf{25.27} & \cellcolor[HTML]{D7FFD7}\textbf{99.29\% }& \cellcolor[HTML]{D7FFD7}\textbf{23.26} & \cellcolor[HTML]{D7FFD7}3\textbf{0.56} & \cellcolor[HTML]{D7FFD7}\textbf{25.34} & \cellcolor[HTML]{D7FFD7}\textbf{27.27} & \cellcolor[HTML]{D7FFD7}\textbf{0.000157} & \cellcolor[HTML]{D7FFD7}\textbf{0.86} \\ \hline
\end{tabular}
}
\caption{\label{tab:widgets}Table showing experiments' results of SIREN, WIRE, the baseline(RELU+POS.ENC.), MFN, and TRIDENT.}
\end{table*}

%% file: sec/5_ablation.tex
\section{Ablation study}
\label{sec:ablation}
Apart from the main experiments in the last section, we also do some additional Ablation experiments to guarantee high representativeness and robustness compared to the latest nonlinear INR. Furthermore, in this session, we also experimented on the choices of hyperparameters. 
\subsection{Choice of Parameter $s_0$}
The parameter $s_0$ controls the length of the window. A too-narrow window reduces representativeness by losing low-frequency information and increasing sensitivity to rapid signal changes. Conversely, a too-large window causes TRIDENT to lose spatial compactness and high-frequency information. Consequently, we conducted a 4X super-resolution experiment with varying $s_0$ values of  1, 5, 10, 20, with the same setting as in section \ref{sr}. The input `pepper' is sourced from the well-known super-resolution dataset  Set14~\cite{zeyde2012single}.

The visualisation results are shown in Figure~\ref{fig:ablation_parameter}. With $s_0=20$, the image quality is poor due to the excessively large window. For $s_0=10$, the result is over-smoothed, losing high-frequency features. However, with $s_0=1$,  the window is too narrow, leading to increased noise sensitivity. Our conclusion is that setting $s_0$ around 5 is optimal, achieving a PSNR of 27.27 and preserving both high-frequency and low-frequency information.
\subsection{Music Audio reconstruction}
To further show the advantage of TRIDENT brought by the nonlinear trilogy: order, frequency, and spatial compactness, we choose a hard task for MLP, music reconstruction. The first 7 seconds from Bach's Cello Suite No.1 is selected as input and the setting is the same as described in Section \ref{audio}. \\
Figure \ref{fig:ablation_audio} shows the visualisation of the reconstruction results of Bach's music via TRIDENT and WIRE. TRIDENT inherits almost all features from the ground truth, while WIRE fails to preserve any of them. Through this ablation experiment, we further show TRIDENT's high representativeness by the nonlinear trilogy.   

\subsection{CT Reconstruction with a decreasing number of measurements}
To test the robustness of TRIDENT in solving inverse problems, we do an experiment on CT Reconstruction with a decreasing number of measurements with the same experiment setting and the same data of \ref{ct}. \\
The CT reconstruction results separately from 100 projections, 75 projections, and 50 projections are visualized in figure \ref{fig:ablation_CT}. In all projections, we hold a leading gap over WIRE, especially in lower projections(SSIM: 0.61(WIRE) vs 0.78(TRIDENT)). TRIDENT is more robust and the leading gap is larger with the dimension decreasing, which enables patients to reduce X-ray exposure.

%% file: sec/6_conclusion.tex
\section{Conclusion}
\label{sec:conclusion}
In this work, we have introduced TRIDENT, a new formula for Implicit Neural Representation, which benefits from the nonlinear trilogy: Order, Frequency, and Spatial Compactness. Strong theoretical principles are provided to show TRIDENT's high representativeness and robustness. Various experiments are conducted to show TRIDENT's versatility and high capacity for solving inverse problems, including CT, audio, 3D occupancy reconstruction, denoising, and super-resolution tasks. TRIDENT achieves always the best and hence it is the best solution for nonlinear INRs.

%% file: sec/7_acknowledgement.tex
\section*{Acknowledgement}
\label{sec:Acknowledgement}
ZS acknowledges support from the Department of Mathematics, College of Science, CityU, and HKASR reaching out award. 
YC and AIAR greatly acknowledge funding from the Cambridge Centre for Data-Driven Discovery and Accelerate Programme for Scientific Discovery, made possible by a donation from Schmidt Futures.
RHC acknowledges support from HKRGC GRF grants CityU1101120 and CityU11309922 and CRF grant C1013-21GF.
AIAR acknowledges support from CMIH (EP/T017961/1) and CCIMI, University of Cambridge. This work was supported in part by Oracle Cloud credits and related resources provided by Oracle for Research. Also, EPSRC Digital Core Capability.
CBS acknowledges support from the Philip Leverhulme Prize, the Royal Society Wolfson Fellowship, the EPSRC advanced career fellowship EP/V029428/1, EPSRC grants EP/S026045/1 and EP/T003553/1, EP/N014588/1, EP/T017961/1, the Wellcome Innovator Awards 215733/Z/19/Z and 221633/Z/20/Z, the European Union Horizon 2020 research and innovation programme under the Marie Skodowska-Curie grant agreement No. 777826 NoMADS, the Cantab Capital Institute for the Mathematics of Information and the Alan Turing Institute.